\documentclass{article}
\usepackage{arxiv}
\usepackage{multirow}
\usepackage{array, tabularx}
\newcolumntype{P}[1]{>{\centering\arraybackslash}p{#1}}
\newcolumntype{C}[1]{>{\raggedright\arraybackslash}p{#1}}
\newcolumntype{M}[1]{>{\centering\arraybackslash}m{#1}}
\usepackage{graphicx}
\usepackage[lofdepth]{subfig}
\usepackage{adjustbox}
\usepackage{tikz}
\usepackage{mathtools}
\usepackage{pdfpages}
\usepackage[numbers,sort&compress]{natbib}

\title{Ensembling Handcrafted Features with Deep Features: An Analytical Study for Classification of Routine Colon Cancer Histopathological Nuclei Images. 
}
\author{
 Suvidha Tripathi \\
  Department of Information Technology\\
  Indian Institute of Information Technology Allahabad\\
  Jhalwa, Deoghat, Prayagraj, Uttar Pradesh 211015 \\
  \texttt{suvitri24@gmail.com} \\
   \And
 Satish Kumar Singh \\
  Department of Information Technology\\
  Indian Institute of Information Technology Allahabad\\
  Jhalwa, Deoghat, Prayagraj, Uttar Pradesh 211015 \\
  \texttt{sk.singh@iiita.ac.in} \\
}
\begin{document}
\maketitle

\begin{abstract}
The use of Deep Learning (DL) based methods in medical \\   histopathology images have been one of the most sought after solutions to classify, segment, and detect diseased biopsy samples. However, given the complex nature of medical datasets due to the presence of intra-class variability and heterogeneity, the use of complex DL models might not give the optimal performance up to the level which is suitable for assisting pathologists. Therefore, ensemble DL methods with the scope of including domain agnostic handcrafted Features (HC-F) inspired this work. We have, through experiments, tried to highlight that a single DL network (domain-specific or state of the art pre-trained models) cannot be directly used as the base model without proper analysis with the relevant dataset. We have used F1-measure, Precision, Recall, AUC, and Cross-Entropy Loss to analyse the performance of our approaches. We observed from the results that the DL features ensemble bring a marked improvement in the overall performance of the model, whereas, domain agnostic HC-F remains dormant on the performance of the DL models.

\end{abstract}

\section{Introduction}

In recent years, there is a surge of deep learning methods in applications of biomedical image analysis. Various problems of detection, segmentation, classification, and compression are solved using deep learning methods in medical datasets. They could achieve a decent level of performance due to the introduction of DL techniques in this domain. However, traditional feature descriptors that provide domain-specific features should not be ignored during the analysis as they have been a widely used feature extraction techniques before Deep Convolutional Neural Networks (DCNNs) gained momentum. Traditional features have an advantage over deep features in terms of feature interpretability and high feature variance even in case of small datasets. In contrast, deep features require lengthy datasets to be able to generalise well on the learned weights. With medical datasets, the availability of large annotated samples is still a challenge and therefore, it imposes a bottleneck for learning very DCNNs. In such scenarios, transfer learning \cite{torrey2009transfer} and fine-tuning is the way forward for generalising medical images for classification tasks. Another advantage of extracting domain-specific feature descriptors is that these features relate well with visual attributes defined by clinicians. Such features can easily be acquired through object level, spatial level, and domain agnostic handcrafted feature descriptors instead of DL methods. Acknowledging the benefits inherited by both handcrafted Features (HC-F) and Deep Learning (DL) features, recently few studies \cite{zhang2018classification,wang2014mitosis} have combined them in different ways to perform classification of medical images. \\
In this paper, through experiments, we delved into the hypothesis that complex problems, especially in context with microscopy histopathology images, could be solved with better performance outcomes by concatenating HC-F and DL features. To justify our hypothesis,  we have performed a detailed analysis of classification performance with HC-F descriptors and DL models. Why DL methods are insufficient for such problems can be understood through our study in \cite{tripathi2018histopathological}. The experimental analysis concluded in this study forms the base of our work.  In \cite{tripathi2018histopathological}, we investigated relatively older deep architectures like AlexNet, VGG16, and VGG19 for their performance on classifying the nuclei dataset. While they gave state of the art performance with ImageNet dataset \cite{russakovsky2015imagenet}, these models could not perform at par with histopathological data. Therefore, we extended our analysis further through this work that DL methods if applied alone, might not perform well; however, their efficiency could be enhanced using simple ensemble models.     
We have mainly chosen the problem of nuclei classification because it is often the ultimate goal in cancer applications. The state or class of nuclei often is a pre-requisite to convey higher-level information about the type of the disease. For example, large, irregular shaped, fragmented, and discoloured nuclei are the signs of the presence of a malignant tumour. Such visual features can be extracted from traditional object level, spatial and graph-based feature descriptor algorithms. However, the main drawback with these descriptors is the requirement for detecting and then segmenting each nucleus from the microscopy image to extract object/spatial/graph-based features \cite{gurcan2009histopathological}. Earlier and recent studies based on these features have performed their classification after segmentation of all the training and testing nuclei \cite{wang2014mitosis,diamond2004use,doyle2007automated,sirinukunwattana2015novel,sirinukunwattana2016locality} 
. However, the segmentation of nuclei comes with its challenges and is still a field of evolving research. Without proper segmentation, there is no guarantee of accurate feature extraction. Therefore, in this study, we have used traditionally handcrafted algorithms that take a raw image as input and calculate features of the object in correlation with its surroundings. The extracted features are then combined with DL features to test whether or not there is a performance gain and whether or not the DL methods, in combination with HC-F, improves the overall performance of the network. 

In summary, the key points of this work are:
\begin{enumerate}
\item Analysis of current deep learning networks for evaluating their performance on complex colon cancer nuclei images.
\item Ensemble HC-F to improve their overall performance.
\item Combining HC-F and DL feature using different methods to study the effect of domain agnostic HC-F over DL features and vice-versa.
\item Contrasting end-to-end deep network performance with transfer learned deep feature classification.
\item We tested different models to thoroughly analyse the effect of using reduced feature sets over complete sets on the individual performance of each model.  
\item We contrasted two different ensemble methods to present an analysis that combining models in a certain way could affect the overall performance. 
\end{enumerate}
\label{intro}
\section{Literature Review}
Deep and handcrafted features quantify image representation in different ways. Handcrafted algorithms take cues from human's interpretation of objects, for example, object's colour, size, surface area, shape, texture, depth, its correlation with its surroundings and ability to distinguish an object from the background. Whereas, DL methods are mainly a pixel-based representation of image features that are learned over ample iterations until the features transform to the closest representation of the image. Most of the earlier traditional feature-based classification and segmentation algorithms derived features from segmented nuclei and glands. For example, object-level features derived from nuclear structures of H\&E stained images yielded 97\% accurate grading of prostate cancer in an experimental study by Jafari-Khouzani and Soltanian-Zadeh \cite{jafari2003multiwavelet}. \cite{sims2003image,gil2002image} reported more OL based feature classification methods where the main focus is on the segmentation of nuclei but also hinted on the usefulness of morphometric and spatial features for improved classification. For quantifying spatial arrangement of cell nuclei and glands into features spatially related features such as Voronoi Tessellation, Delaunay Triangulation, Minimum Spanning Tree, and Active Contours are some of the graph-based structures widely used in previous pieces of literature \cite{demir2005automated,weyn1998automated,keenan2000automated,dhandra2006endoscopic}. The drawback with object and spatial level features is their requirement for segmentation of the object before calculating the features. 
\par In our work, we did not segment the nuclei from the background features and used the raw images for training the proposed ensemble models. Some of the recent studies that have used raw images for extracting features from medical datasets have used mainly deep learning algorithms; like a pre-trained CNN can be used to extract features for transfer learning. Some studies have claimed that models pre-trained on independent datasets do not give quality features and instead, models pre-trained on more massive datasets but fine-tuned on a smaller, and related dataset produces higher performance metrics. For instance, Gao et al. \cite{gao2016hep} used deep CNN to classify human epithelial-2 images and found out that fine-tuned pre-trained nets gave higher accuracy than those trained from scratch on a smaller dataset. This strategy has also been verified in \cite{tajbakhsh2016convolutional}. Since 2015, numerous studies have published the benefits of deep learning-based classifiers for classification of cell \cite{chen2016deep}, brain tumour histopathology images \cite{xu2015deep}, to estimate the number and proportion of Micro Circulatory Supply Units (MCSU) in human squamous cell carcinoma by Carneiro et al. \cite{carneiro2015weakly}. Latest CAMELYON 2016 \cite{bejnordi2017diagnostic} and 2017 challenge \cite{litjens20181399} has seen the widespread use of DL methods in breast cancer and Sentinel Lymph node metastases detection and classification. These deep learning-based studies have although used raw images as input to their model but used some pre-processing methods. Such as the authors in \cite{chen2016deep}, calculated morphological, optical phase, and optical loss features from the detected cells and then used deep neural networks for classifying these features. 
\par There are also recent studies on the use of both HC-F and DL features. Wang et al. \cite{wang2014mitosis} used domain inspired HC-F in combination with generalised DL features for mitosis detection in breast cancer pathology images. They proposed a two-stage cascaded ensemble technique where they first classified mitoses using HC-F and DL features separately and then in the second stage, combined them for improved performance. Their methodology was, however, based on object-level features like morphology, intensity, and texture, which require candidate segmentation before extracting features. Similarly, authors in \cite{zhang2018classification} claimed that combining BoF and LBP descriptors with deep VGGf and Caffe-ref features with reduced dimensions enhanced their classification accuracy. It is the only study based on handcrafted descriptors that work on raw image data rather than first segmenting the object of interest. However, their method cannot be evaluated based on accuracy alone since this is not the accepted metric for evaluating an unbalanced dataset. Moreover, their analysis lacks detailed discussion as to why HC-F enhanced performance. They used relatively old deep networks for justifying their claim when there are more new networks that may achieve comparable performance without the requirement of handcrafted features. \
All these studies have paved the scope for more open research possibilities of combining traditional HC-F with deep features where one approach inspires domain-specific responses; the other approach increases the generalising capability. But, none of these works have analysed and justified the claim that combining both approaches curb the individual limitations posed by each method. Therefore, through our work, we have tried to overcome the drawbacks of each of these pieces of literature by \begin{itemize}
\item Using raw images as input without the need to segment the candidate nuclei.
\item Highlighting the need for extensive model selection experiments to identify the best performing method using the ensemble techniques.
\item Extensive analysis to justify the use of ensemble methods over single deep learning methods.
\item Using more relevant performance metrics to validate our results 
\end{itemize}

\label{litrev}
\section{Methodology}

\begin{table}[htbp]
\caption{Summary of deep architectures used in the study}
\label{tab1}       
\centering
\begin{adjustbox}{width=\textwidth}
\renewcommand{\arraystretch}{1.5}
\begin{tabular}{|P{23mm}|P{16mm}|P{13mm}|P{19mm}|P{14mm}|P{14mm}|} 
\noalign{\smallskip} \hline
Architecture & No. of Conv \newline layers & No. of FC \newline layers & No. of training \newline  parameters & Minimum image size & top 5 error \newline on ImageNet \\
\hline
AlexNet \cite{szegedy2015going} & 5 & 3 & 56 millions &$224 \times 224$ & 15.3\% \\
\hline
VGG16 \cite{simonyan2014very} & 13 & 3 & 134 millions & $227 \times 227$ & 7.3\% \\
\hline 
VGG19 \cite{simonyan2014very} &16&3&139 millions&$227 \times 227$ & 8.0\%\\
\hline 
ResNet50 \cite{he2016deep} &48& 1&23 millions &$197 \times 197$ & 3.57\%\\
\hline
DenseNet121 \cite{huang2017densely} &120&1&7 millions& $221 \times 221$ & 7.71\% \\
\hline
InceptionV3 \cite{szegedy2015rethinking} &42&1&22 millions& $299 \times 299$ & 3.08\% \\
\hline \noalign{\smallskip}
\end{tabular}
\end{adjustbox}
\end{table}
\subsection{DL Features Extraction}
\label{DL}
We fine-tuned both early and recent state of the art DCNNs pre-trained on ImageNet dataset \cite{russakovsky2015imagenet} with our dataset and then extracted features from the last Fully Connected (FC) layers of each network. Table \ref{tab1} summarizes the details about their layers, number of training parameters, minimum image size required as input, and top 5 error achieved on ImageNet database by each architecture. For training these networks mentioned in Table \ref{tab1}, we used pre-trained network weights and finely tuned them on our dataset for our four-category nuclei classification. For four class classification, we modified 1000 neurons to 4 neurons in the last FC layer of the original network. Later, to adapt to the input size requirements of each network (Table \ref{tab1}), we resized our images to the minimum image input size. For each case, the maximum number of epochs were set to 100 and we used stochastic gradient descent with momentum as our optimizer. The learning rate was optimized at 0.001 and kept constant throughout the network training. The momentum value was set to 0.95 for all training networks. To monitor the progress, we randomly chose 15\% of the data as our validation set, 70\% of the data for training and remaining 15\% for testing.
\par After training each network, we saved the output weights of the FC layer of each network. The obtained output are the learned feature maps from the last deep learning layer (convolutional or pooling layer) mapped to the number of neurons or nodes of the following FC layer. These number of nodes vary with the architecture. For example, AlexNet uses 3 fully connected layers with different numbers of nodes in each of these layers, and we extracted features from the first fully connected layer having 4096 nodes.
Similarly, for ResNet50, the single fully connected layer has 2048 nodes, hence the features per image sample has dimension $1 \times 2048$.  Table \ref{tab2} describes the number of features extracted for each deep architecture. The features obtained after the last convolutional layer describe the most sophisticated characteristics of the sample into numerical values. The data analysis of these features is out of the scope of this work. 

\begin{table}[htbp]
\caption{Architectures and number of features per image}
\label{tab2}       
\centering
\renewcommand{\arraystretch}{1.5}
\begin{tabular}{|P{23mm}|P{16mm}|} 
\noalign{\smallskip} \hline
Architecture & No. of features \\
\hline
AlexNet \cite{szegedy2015going} & 4096 \\
\hline
VGG16 \cite{simonyan2014very} & 4096 \\
\hline 
VGG19 \cite{simonyan2014very} & 4096\\
\hline 
ResNet50 \cite{he2016deep} & 2048\\
\hline
DenseNet121 \cite{huang2017densely} & 1024\\
\hline
InceptionV3 \cite{szegedy2015rethinking} & 2048 \\
\hline \noalign{\smallskip}
\end{tabular}
\end{table}

\subsection{Handcrafted Feature Extraction}
\label{HF}
\subsubsection{Features}
We used several handcrafted algorithms briefly described in Table \ref{tab3} to quantify each image into features.

\begin{table}[htbp]
\caption{Handcrafted algorithms in summary}
\label{tab3}       
\renewcommand{\arraystretch}{1.5}
\begin{adjustbox}{width=\textwidth}
\begin{tabular}{|M{20mm}|C{100mm}|M{16mm}|} 
\noalign{\smallskip} \hline
\textbf{Algorithm} & \centering\textbf{Brief summary} & \textbf{No. of features} \\
\hline
\textbf{HoG} \cite{marshall1989method} &Histogram of Oriented Gradients (HoG) looks for the occurrences of intensity gradients and edge directions within an image by calculating for each pixel. The bins with similar orientation create a histogram for an image. This histogram is quantified into features. The key advantage of this method is that it normalises each block of an image locally before histogram calculation for improved performance. & 256\\
\hline
\textbf{LBP} \cite{ojala1996comparative} &Local Binary Patterns (LBP) is widely used feature descriptors for encoding texture information. The four-step process starts with calculating the intensity difference between the centre and neighbouring pixels at radius R. All the negative difference values are converted to zero and positives to ones. It forms a binary pattern concerning the centre pixel. The centre pixel is then replaced by the decimal equivalent of the binary pattern. &59 \\
\hline 
\textbf{BoVW} \cite{fei2005bayesian} & Bag of Visual Words (BoVW) encode images into key points instead of words and form a vector of keypoint descriptors extracted from the image. The process involves extracting features followed by k-means clustering. Each cluster centre represents a feature or visual word. A vocabulary of these visual words contains the cluster centre features, and the number of clusters defined at the start gives the number of features per image. This method has an advantage over other handcrafted algorithms because it does not require localisation for the detection of objects.& 100 (in our work)\\
\hline 
\textbf{SURF} \cite{bay2008speeded} & Speeded-Up Robust Features (SURF) is a widely used keypoint detector descriptor enhanced by its robustness due to its scale and rotation invariance properties. SURF is much faster in comparison to other proposed keypoint detectors like Hessian Matrix-based measure for the detector and a distribution-based descriptor. Due to its 2D variable feature-length per image, we have reshaped each image descriptor into 1D and padded the remaining columns with 0 to obtain a uniform dimension matrix from the whole dataset. & 1216\\
\hline
\textbf{LDEP} \cite{dubey2015local2} & Local Diagonal Extrema Pattern (LDEP) as the name suggests, it encodes both, the relationship among neighbours and the relationship among neighbouring pixels with the centre. However, only diagonal neighbours are considered for encoding a descriptor which greatly reduces the dimension of the feature vector while achieving the gain in speed. LDEP also finds its applications in CT image databases. &24\\
\hline
\textbf{LWP} \cite{dubey2015local1} &Local Wavelet Pattern (LWP) does not ignore the inter-neighbour relationship. It encodes this information using 1-D Haar wavelet decomposition. It has its applications in medical CT databases. &256 \\
\hline
\textbf{LCOD} \cite{dubey2015local4} &Local Colour Occurrence Descriptor (LCOD) quantises RGB colour channels of coloured images into few shades for dimension reduction and as well for forming an encoded pattern in the form of a descriptor. The occurrence of each quantised colour in the image gives a hint about the homogeneous regions. This descriptor has an advantage over LBP and LBDP as it is rotation, scale, and illumination independent. &256 \\
\hline
\textbf{RSHD} \cite{dubey2015rotation} & Rotation and Scale-invariant Hybrid image Descriptor (RSHD). Besides LCOD and BoVW, all other descriptors do not fair well with coloured images. Also, most of them are sensitive to rotation and scale variations. RSHD is a hybrid descriptor that encodes an RGB colour palette into 64 shades. Along with colour, the texture is also extracted using 5 rotation invariant structuring elements. Finally, colour and texture information are fused to form an RSHD feature descriptor. &320 \\
\hline
\textbf{LBDP} \cite{dubey2015local3} & In the original literature, Local Bit-plane Decoded Pattern (LBDP) has been used to encode MRI and CT databases into descriptors for retrieval purposes. With the same dimension as LBP, LBDP not only encodes the relationship of the centre pixel with its neighbours but also among local neighbours in each bit plane.  &256 \\
\hline \noalign{\smallskip}
\end{tabular}
\end{adjustbox}
\end{table}

The main advantage of these HC-F is their ability to transform raw images into feature descriptors without going through the process of segmentation. 

\subsubsection{Feature Selection}
\label{PCA}
From Table \ref{tab3}, we can see the number of features for each descriptor algorithm. To select the best set of features with high variance or the features that capture most information of the data, we used PCA \cite{pearson1901liii} and selected the top 100 component variations in the case of handcrafted features and top 1000 in the case of deep learning features. Since the variance signifies the discriminative capacity of the feature vectors, the respective features are selected based on decreasing variance. Hence, most discriminative features projected on PCA space has been selected to enhance the intra-class similarity and diminish the inter-class similarity.   

\subsubsection{Classifiers}
\label{classifier}
\begin{figure}
\includegraphics[width=\textwidth, height=3in]{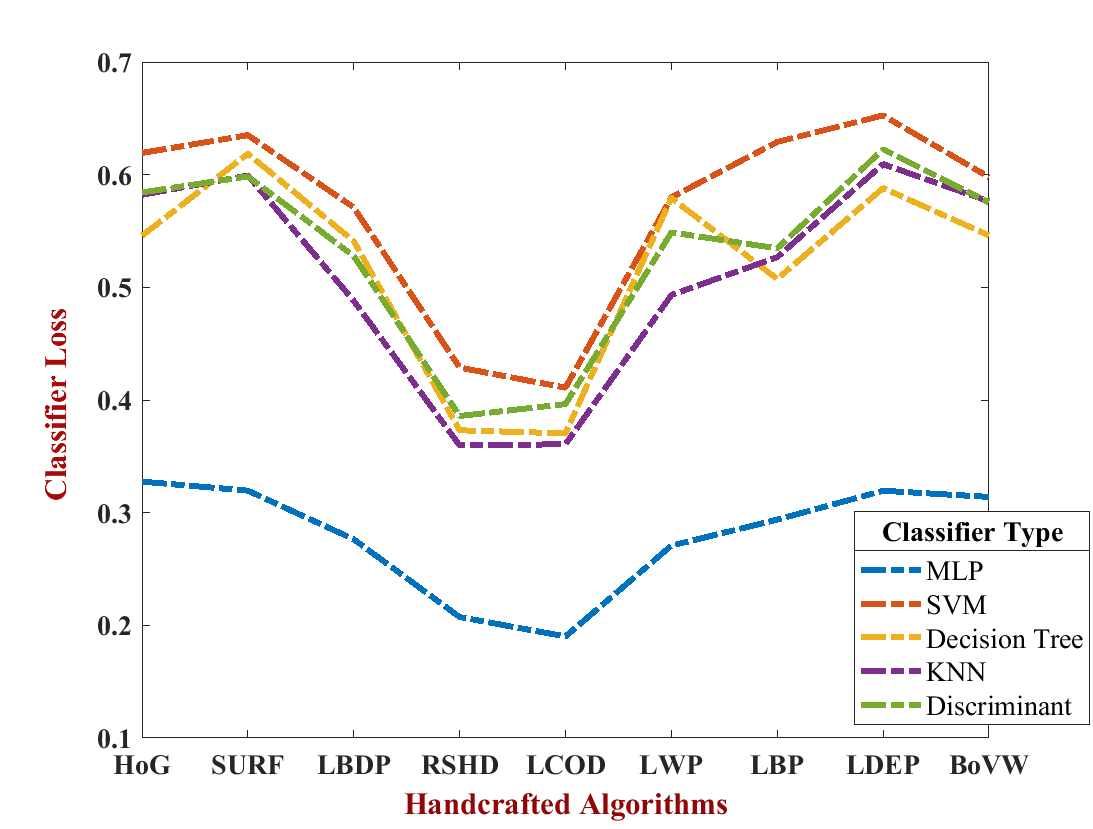}
\caption{Classifier v/s Loss curve}
\label{fig1}       
\end{figure}
For classifying nuclei with HC-F, we tested each descriptor with five classifiers: Multi-Layer Perceptron (MLP), SVM, Decision trees, KNNs, and Kernel Discriminant model. The following graph in Figure \ref{fig1} shows the loss values we obtained for each algorithm. The main aim of this experiment was to decide the type of classifier to be used to discriminate the features into classes. For this, one of the five classifiers was trained with the training set features obtained from each handcrafted algorithm (HoG, SURF, LBDP, RSHD, LCOD, LWP, LBP, LDEP, BoVW). We used 2 fold cross-validation to verify the trained model performance and then for the final step, tested the classifier with our test set features. We recorded the loss obtained against the classifier and plotted it as Y-axis values against the respective handcrafted algorithm as X-axis values. The process is repeated for all the classifiers. We could observe from the figure that the curves are not smooth since nine discreet loss values have been plotted for each classifier type and then joined together to form a curve. 
From Figure \ref{fig1}, we observe that the loss for MLP is the least across all the handcrafted features, uniformly. Whereas, the loss curves for other classifiers have crossed each other, indicating that while the loss is less for some algorithms, it may increase for another handcrafted algorithm. This uncertainty about the behaviour of the classifier might have affected the performance of the final ensemble model. Therefore, we used MLP as our final classifier for classifying all handcrafted features and as well the final ensemble model. MLP has one hidden layer with 10 nodes, and the classification layer has 4 output nodes. 
\subsection{Ensemble methods}
We proposed two types of ensemble methods for analysing the effect of handcrafted algorithms on deep learning networks. We performed a series of experiments to produce our final observations. The aim was to capture easily implementable scenarios for the fusion of two domain features. Before the start of the experiment, we assumed that the handcrafted features of the colon cell nuclei might have three possible effects on the deep learning features, positive, neutral, or negative. Therefore, these ensemble methods were formulated for the analysis of the obtained classification results. The complete architecture is described in the following subsections.
\subsubsection{Cascaded Ensemble}
\label{CE}
\begin{figure}
  \includegraphics[width=\textwidth, height=3in]{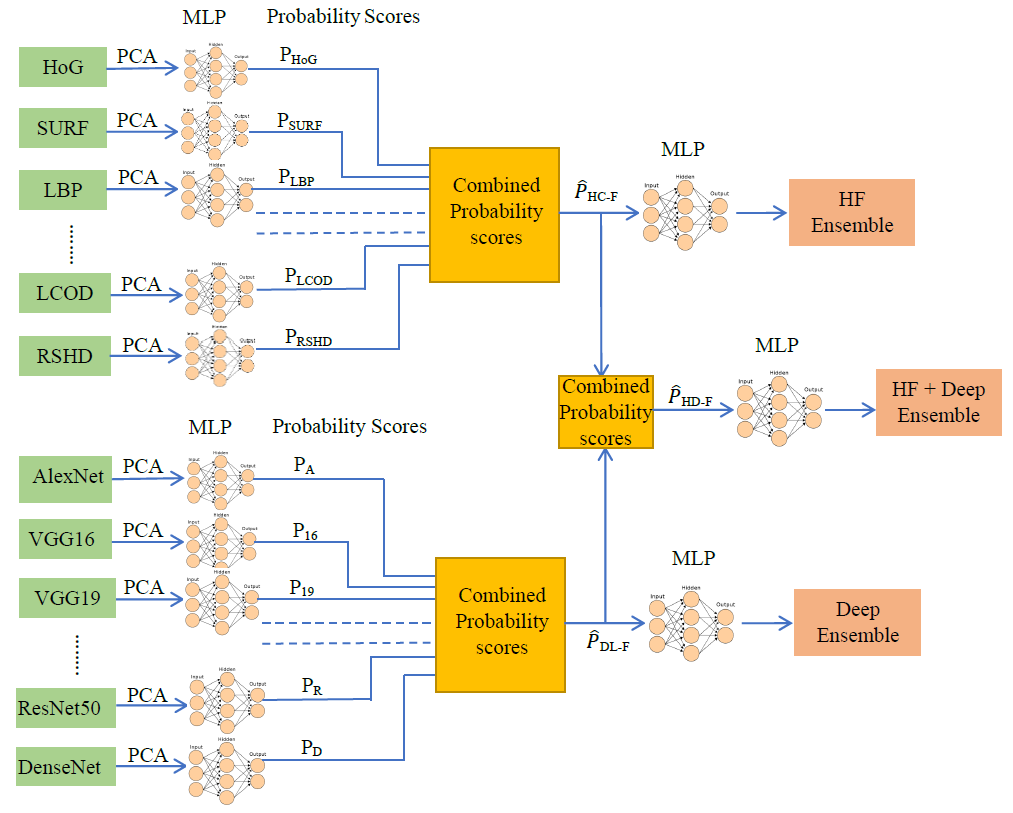}
\caption{Workflow of cascaded ensemble which comprises 3 stages. First, ensemble handcrafted features classification. Second, ensemble deep feature classification. Finally, the third and the final stage, ensemble handcrafted and deep feature classification.}
\label{fig2}       
\end{figure}

The cascaded ensemble produces three classification outputs using the pipeline illustrated in figure \ref{fig2}. First, the original feature matrix from each handcrafted and deep learning algorithms is obtained following the processes described in section \ref{DL} and \ref{HF}, respectively. The feature sets were then reduced using PCA as described in section \ref{PCA}. These reduced feature matrices were then separately classified using the selected MLP classifier. The selection of the classifier was made through the process described in section \ref{classifier}. Each algorithm produced $number\_of\_samples \times 4$ probability vector after MLP classification, where 4 are the number of classes. Once the probability scores from each tested algorithm (handcrafted or deep) are calculated, we constructed the model to give 3 separate outputs.\\
First, all the handcrafted algorithms probability scores were combined using the formula described in equation \ref{eq1} \cite{genest1986combining}. if \textit{$P_{HoG}$, $P_{SURF}$, $P_{LBP}$, $P_{LBDP}$, $P_{LCOD}$, $P_{RSHD}$, $P_{LDEP}$, $P_{LWP}$, and $P_{LDEP}$} are the probabilities obtained from nine handcrafted algorithms then their respective probability scores were combined to obtain final scores ($\hat{P}_{HC-F}$) of ensemble handcrafted features. 
\begin{equation}
\hat{P}=\frac{\prod^N_{i=1} p_i^{w_i}}{\prod^N_{i=1} p_i^{w_i} + \prod^N_{i=1} (1-p_i)^{w_i}}
\label{eq1}
\end{equation}                                                                                                                                                                                                                                                                                                                                                                                                                                                                                                                                                                                                                                                                                                                                                                                                                                                                                                                                                                                                                                                                                                                                                                                                                                                                                                                                                                                                                                                                                                                         
where, $w_i$ are weights and mostly $w_i=\frac{1}{N}$ when a priori information about class probabilities doesn't exists. Here $N$ is equal to the number of methods being aggregated and $p_i$ is the probability of the ith method. We aggregate $\hat{P}$ for all classes and then classify using aggregated probabilities as our input to the MLP classifier to obtain the final ensemble output. The complete process illustrates the "HF- Ensemble" output, as shown in Fig. \ref{fig2}. \\
Second, the probabilities ($P_{A}$ - AlexNet, $P_{16}$ - VGG16, $P_{19}$ - VGG19, $P_{I}$ - InceptionV3, $P_{R}$ - ResNet50, $P_{D}$ - DenseNet121) obtained from deep features classification is combined using the same formula to obtain a combined scores  $\hat{P}_{DL-F}$ of ensemble deep features.  $\hat{P}_{DL-F}$ is taken as a new set of features that were then classified using MLP to obtain final probability scores that describe the "Deep Ensemble" output or the classification model for the deep ensemble.  \\
Third, the combined scores $\hat{P}_{HC-F}$ and $\hat{P}_{DL-F}$ were once again combined using equation \ref{eq1} to obtain $\hat{P}_{HD-F}$ or a combined score of handcrafted and deep feature classification. $\hat{P}_{HD-F}$ when classified through non-linear classifier (MLP) produced "HF + Deep Ensemble" classification probabilities. 

\subsubsection{Feature Concatenation}
\label{FC}
\begin{figure}
  \includegraphics[width=\textwidth, height=2in]{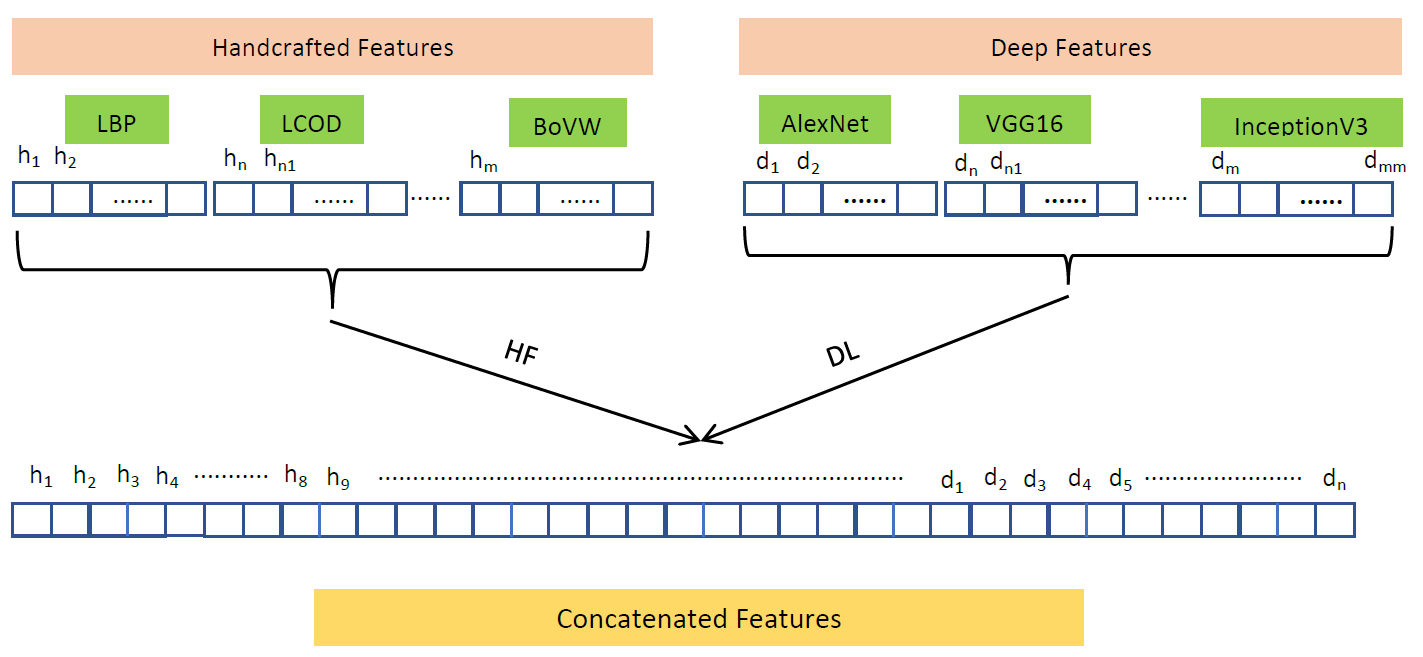}
\caption{Illustration of concatenating different feature sets}
\label{fig3}       
\end{figure}

Besides cascading HC-F and DL features, another method we used to fuse respective feature sets is concatenating them along the length of features. Figure \ref{fig3} illustrates the process. In this method, first, the handcrafted features (HC-F) from individual algorithms are concatenated along the feature dimensions. For example, let $N \times M_1$, $N \times M_2$, $N \times M_3$ are the dimensions of feature matrix obtained from algorithms $X_1$, $X_2$, and $X_3$ where, $N$ is the number of samples and $M_i$ is the feature dimension obtained from the algorithm $i$, where $i=1,2.3\ldots,number\_of\_algorithms$ . Then, the final feature matrix obtained from the concatenation method would be $N \times (M_1+M_2+M_3)$. Individual DL features are concatenated using a similar process. In table 2, feature dimensions obtained from AlexNet, ResNet50, DenseNet121 are 4096, 2048, and 1024, respectively. The number of samples in our work is 29,771. So, the concatenated matrix would be $ 29771 \times 4096+2048+1024 = 29771 \times 7168$. In the final step, the concatenated set of HC-F and DL features are concatenated along the columns or second dimension while the number of rows remains the same (number of samples). This method appends the different features from different algorithms along the single dimension and does not require any pre-processing steps to refine the concatenating feature sets.
\label{method}
\section{Experimental Results}
\subsection{Dataset}
\begin{figure}
\centering
  \includegraphics[width=0.8\textwidth, height=2in]{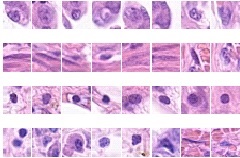}
\caption{Example samples of each class. From first row to fourth: Epithelial Nuclei, Fibroblast Nuclei, Inflammatory Nuclei, and Miscellaneous.}
\label{nuclei}       
\end{figure}

The dataset from \cite{sirinukunwattana2016locality} called as "CRCHistoPhenotypes" consists of 100 H\&E stained microscopy images of $500\times 500$ pixel dimensions each. 100 images have different types of nuclei whose centre coordinate location has been provided as ground truth. The nuclei are divided into 4 categories or classes, Epithelial, Fibroblast, Inflammatory, and miscellaneous. The miscellaneous category includes mitotic figures, adipocyte, endothelial nucleus, and necrotic nucleus. Figure \ref{nuclei} shows the example of nuclei samples from the dataset. We used centre pixel coordinates of nuclei and extracted a $27\times 27$ window around nuclei centres. The final nuclei dataset after collecting nuclei from 100 microscopy images contains 7722 Epithelial, 5712 Fibroblasts, 6970 Inflammatory, and 2039 miscellaneous nuclei. Since the dataset is unbalanced, we used the Adaptive Synthetic Sampling technique \cite{he2008adasyn} for each class to accumulate unbalanced class samples. After balancing, the class count increased to 7722, 7275, 6970, 7804 samples for Epithelial, Fibroblast, Inflammatory, and miscellaneous, respectively. 

\subsection{Performance Measures}
We calculated average $Precision$, $Recall$, $F1$, multiclass $AUC$, and $cross-entropy$ loss values to measure the performance of our methods. The following equations describe the performance measures where $TP$ are the number of True Positives samples, $FP$ is the number of False Positive samples and False Negative ($FN$) describe the number of samples that are positive but predicted as negatives. Similarly, $FP$ or False positives are the samples that are negative but predicted as positives. 
\begin{equation}
Recall= \frac{TP}{TP+FN}
\end{equation} 
\begin{equation}
Precision= \frac{TP}{TP+FP}
\end{equation} 
\begin{equation}
F1 = \frac{2\times Precision\times Recall}{Precision+Recall}
\end{equation} 
\begin{equation}
Loss= -\sum_{c=1}^My_{o,c}\log(p_{o,c})
\end{equation} 

where, $M$ - number of classes, log - the natural log, $y$ - binary indicator (0 or 1) if class label $c$ is the correct classification for observation $o$, $p$ - predicted probability observation $o$ is of class $c$

\subsection{Results}
\label{results}
We have divided our analysis into the following categories:
\begin{itemize}
\item Classification with reduced feature sets (with PCA)
\item Classification with full feature sets (without PCA)
\end{itemize}
Each of the categories is further subdivided into the following experiments
\begin{enumerate}
\item Only handcrafted feature classification
\item Only deep learning features
\item Ensemble of handcrafted feature classification
\item Ensemble of deep feature classification
\item Ensemble of deep + handcrafted features
\end{enumerate}
We did ensembling in two ways:
\begin{itemize}
\item Cascaded Ensemble (described in Section \ref{CE})
\item Concatenation of features (described in Section \ref{FC})
\end{itemize}

The Table \ref{tab4} lists all the handcrafted Algorithms performance with full feature length with respect to colon nuclei classification. We observed that among all the nine methods considered, LCOD and RSHD performed relatively better. The reason being LCOD and RSHD encode colors  and are rotation and scale invariant. H\&E stained histopathology images are highly color sensitive due to the presence of some kind of proteins that react to staining dye by producing dark blue or purplish hue for nuclei and pink color for cytoplasm in the cell. Whereas, all other handcrafted algorithms accept greyscale image as input for further processing which in turn cause loss of critical information in the case of color sensitive histopathology images.
\begin{table}[htbp]
\caption{Comparative performance measures for HC-F classification (without PCA). The best performing metric is marked in bold}
\label{tab4}      
\centering
\renewcommand{\arraystretch}{1.5}
\begin{tabular}{|P{10mm}|P{15mm}|P{10mm}|P{10mm}|P{10mm}|P{10mm}|} 
\noalign{\smallskip} \hline
\textbf{Method} & \textbf{Precision} & \textbf{Recall} & \textbf{F1-Score} & \textbf{AUC} & \textbf{Loss} \\
\hline
HoG &0.4176  &0.4172 &0.4150  &0.6803 &0.3233  \\
\hline
LBP & 0.4718&0.4713 &0.4667 &0.7308 &0.2986  \\
\hline 
BoVW &0.4056&0.4087&0.4042&0.6768&0.3173 \\
\hline 
SURF &0.3719&0.3789&0.3614&0.6548&0.3246 \\
\hline
LBDP &0.5372&0.5454&0.5391&0.7986&0.2594 \\
\hline
LDEP &0.4014&0.3988&0.3927&0.6628&0.3201  \\
\hline
LWP &0.5169&0.5168&0.5168&0.7763&0.2773  \\
\hline
\textbf{RSHD} &\textbf{0.6596}&\textbf{0.6629}&\textbf{0.6606}&\textbf{0.8798}&\textbf{0.2025}  \\
\hline
LCOD &0.6483&0.6517&0.6492&0.8730&0.2084  \\
\hline \noalign{\smallskip}
\end{tabular}
\end{table}
The Table \ref{tab5} describe deep networks performance fine tuned on balanced nuclei dataset. The output we obtain is end to end classification by softmax layer of the networks.   From the results, we observe that ResNet50 performed better than all other networks. Though the difference in performance is clearly visible which is within ~7\% range but, the better comparative metric here should be the number of parameters trained and the time taken to train each network. Any network with minimum space and time complexity can be optimized for classification task. However, the 7\% difference could become large depending on the application. When using a single deep architecture trained or fine-tuned on nuclei dataset,  it is always better to use highest performing architecture.     
\begin{table}[htbp]
\caption{Comparative performance measures for fine tuned Deep networks. The best performing metric is marked in bold}
\label{tab5}      
\centering
\renewcommand{\arraystretch}{1.5}
\begin{tabular}{|P{23mm}|P{15mm}|P{10mm}|P{10mm}|P{10mm}|P{10mm}|} 
\noalign{\smallskip} \hline
\textbf{Method} & \textbf{Precision} & \textbf{Recall} & \textbf{F1-Score} & \textbf{AUC} & \textbf{Loss} \\
\hline
AlexNet & 0.8280 & 0.8281&0.8276  &0.9386 &1.1469  \\
\hline
VGG16 & 0.8693& 0.8699& 0.8689&0.9744 &0.8372  \\
\hline 
VGG19 &0.8575&0.8581&0.8578&0.9701&0.9222 \\
\hline 
InceptionV3 &0.8164&0.8175&0.8175&0.9538&1.0795 \\
\hline
\textbf{ResNet50} &\textbf{0.8900}&\textbf{0.8893}&\textbf{0.8892}&\textbf{0.9799}&\textbf{0.7335} \\
\hline
DenseNet121 &0.8784&0.8706&0.8706&0.9756&0.7972  \\
\hline \noalign{\smallskip}
\end{tabular}
\end{table} 

The table \ref{tab6} compares the performance of each deep model for their quality of features they output and the consequential classification performance of those features when passed through layers of Multi-Layer Perceptron (MLP). The metrics in Table \ref{tab6} suggests that transfer learning, in this case, has proved to show a considerable boost in performance for all networks but VGG19. In the literature of VGG19 \cite{simonyan2014very}, the authors stopped adding layers to VGG16 since no improvement was recorded when more layers were added. Similarly, this behaviour was again observed with our dataset as evident from the results. Observed performance measures for each model suggests that the extracted features when transfer learned with one or more layers of MLP network perform better than deep end-to-end architectures.   

\begin{table}[htbp]
\caption{Comparative performance measures for Deep features after MLP classification. The best performing metric is marked in bold}
\label{tab6}      
\centering
\renewcommand{\arraystretch}{1.5}
\begin{tabular}{|P{23mm}|P{15mm}|P{10mm}|P{10mm}|P{10mm}|P{10mm}|} 
\noalign{\smallskip} \hline
\textbf{Method} & \textbf{Precision} & \textbf{Recall} & \textbf{F1-Score} & \textbf{AUC} & \textbf{Loss} \\
\hline
AlexNet &0.9560  &0.9561 &0.9560  &0.9949 &0.0355  \\
\hline
VGG16 & 0.9596 &0.9596 &0.9596 &0.9957 &0.0326  \\
\hline 
VGG19 &0.8645&0.8638&0.8640&0.9744&0.0920 \\
\hline 
InceptionV3 &0.9583&0.9584&0.9583&0.9962&0.0304 \\
\hline
\textbf{ResNet50} &\textbf{0.9651}&\textbf{0.9652}&\textbf{0.9651}&\textbf{0.9968}&\textbf{0.0273} \\
\hline
DenseNet121 &0.9624&0.9626&0.9625&0.9964&0.0308  \\
\hline \noalign{\smallskip}
\end{tabular}
\end{table}

The observations of the Cascaded Ensemble described in Section \ref{CE} are recorded in Table \ref{tab7}. The three outputs are ensemble output of handcrafted features, ensemble output of deep features, combined ensemble output of handcrafted and deep features, mentioned in the table in the same order. To validate our ensemble model, we performed the experiments with two, five, and ten folds cross-validation. We observed that the performance difference between deep ensemble and HC-F + DL ensemble for Precision is 0.62\%, Recall is 0.63\%, F1 is 0.62\%, AUC is 0.03\%, and loss is 0.0043. The values suggested that the difference is small and the performance could be considered similar. The range of difference falls after two decimal points which may or may not be significant for certain applications.  Section \ref{discussion} discusses the probable reasons.

\begin{table}[htbp]
\caption{Comparative performance measures for cascaded ensemble (ensemble method 1) validated at different cross-validation folds. The best performing metric is marked in bold}
\label{tab7}      
\centering
\renewcommand{\arraystretch}{1.5}
\begin{tabular}{|P{23mm}|P{15mm}|P{10mm}|P{10mm}|P{12mm}|} 
\noalign{\smallskip} \hline
\textbf{Method} & \textbf{Metrics} & \textbf{2-folds} & \textbf{5-folds} & \textbf{10-folds}\\
\hline
\multirow{5}{*}{HC-F ensemble} & Precision &0.7204&0.7212&0.7204\\
\cline{3-5}
&Recall&0.7211&0.7219&0.7214\\
\cline{3-5}
&F1&0.7202&0.7209&0.7202\\
\cline{3-5}
&AUC&0.9190&0.9125&0.9132\\
\cline{3-5}
&Loss&0.1785&0.1771&0.1764\\
\hline
\multirow{5}{*}{\textbf{Deep ensemble}} &Precision&0.9848&\textbf{0.9873}&0.9872\\
\cline{3-5}
&Recall&0.9853&\textbf{0.9876}&\textbf{0.9876}\\
\cline{3-5}
&F1&0.9850&0.9873&\textbf{0.9874}\\
\cline{3-5}
&AUC&0.9974&\textbf{0.9981}&\textbf{0.9981}\\
\cline{3-5}
&Loss&0.0155&\textbf{0.0130}&\textbf{0.0130}\\
\hline 
\multirow{5}{*}{\shortstack{HC-F + DL\\ ensemble}} &Precision&0.9789&0.9809&0.9811\\
\cline{3-5}
&Recall&0.9793&0.9811&0.9813\\
\cline{3-5}
&F1&0.9790&0.9810&0.9812\\
\cline{3-5}
&AUC&0.9971&0.9978&0.9978\\
\cline{3-5}
&Loss&0.0197&0.0174&0.0173\\
\hline \noalign{\smallskip}
\end{tabular}
\end{table}

Table \ref{tab8} records the performance of the feature concatenation method using the process described in Section \ref{FC}. To highlight whether reducing features have had any effect on the performance of the model or not we have segregated the results in two ways: observations with reduced features (PCA) and with original features (Without PCA). 'Handcrafted (HC-F)' is a concatenation of all handcrafted features, 'Deep (Deep-F)' is concatenation of all deep features, 'HC-F + Deep-F' is concatenation of all handcrafted and deep features, 'HC-F +ResNet50' concatenates all handcrafted features and ResNet50 features only, similarly for 'HC-F + DenseNet121' and 'HC-F + InceptionV3'. We observed that the performance metrics tend to behave differently with the number of validation folds in 'HC-F', 'Deep-F', and 'HC-F + Deep-F'. When 2 fold cross-validation was used to validate the model, the metrics obtained is higher than the metrics obtained in the case of 5 and 10 folds cross-validation. We observed the trend irrespective of whether the models have been validated with or without PCA reduced features. However, a combination of handcrafted features with only one deep learning model (ResNet50, DenseNet121, or InceptionV3), the observed values between folds is similar relative to other models under consideration in the table \ref{tab8}. Following the obtained observations, we decided to reject 2-fold cross-validation metrics in the case of this ensemble model. We also observed that 'Deep-F' and 'HC-F + Deep-F' models yielded a performance decline when they used features without PCA reduction (considering only 5 and 10 folds cross-validated metrics). Whereas, all the other models have shown an increase in performance. For 10-fold cross-validation, 'HC-F+ResNet50' without PCA performed best among all models. At the same time, we observed no marked difference in performance metrics of 'Deep-F' and 'HC-F + Deep-F' indicating that no additional discriminating HC-F features influenced the performance of the deep features. Compared to the first ensemble method (cascaded ensemble), this ensemble method could not enhance the performance of 'Deep-F' or 'HC-F + Deep-F'. 
\begin{table}[htbp]
\caption{Comparative performance measures for concatenating features (ensemble method 2) validated at different cross-validation folds.}
\label{tab8}      
\centering
\begin{adjustbox}{width=\textwidth}
\renewcommand{\arraystretch}{1.7}
\begin{tabular}{|P{23mm}|P{20mm}|P{10mm}|P{10mm}|P{10mm}|P{10mm}|P{10mm}|P{10mm}|} 
\noalign{\smallskip} \hline
\multirow{2}{*}{\textbf{Method}} & &\multicolumn{3}{c|}{\textbf{PCA}} & \multicolumn{3}{c|}{\textbf{Without PCA}}\\
\cline{2-8}
&\tikz{\node[below left, inner sep=0.5pt] (def) {Metrics};%
      \node[above right,inner sep=0.5pt] (abc) {Folds};%
      \draw (def.north west|-abc.north west) -- (def.south east-|abc.south east);}
&2&5&10&2&5&10 \\
\hline
\multirow{4}{*}{\shortstack{handcrafted\\ (HC-F)}}&F1&0.6755&0.6937&0.6868&0.6732&0.6911&0.6740\\
\cline{3-8}
&AUC&0.8865&0.8669&0.8956&0.8893&0.8975&0.8866\\
\cline{3-8}
&Loss&0.0100&0.0095&0.0096&0.0098&0.0095&0.0100\\
\cline{3-8}
&No. of Features&\multicolumn{3}{M{30mm}|}{783}&\multicolumn{3}{M{30mm}|}{2631 }\\
\hline
\multirow{4}{*}{Deep (Deep-F)}&F1&0.9879&0.9569&0.9577&0.9867&0.9256&0.9260\\
\cline{3-8}
&AUC&0.9995&0.9965&0.9966&0.9994&0.9915&0.9913\\
\cline{3-8}
&Loss&0.0005&0.0016&0.0015&0.0005&0.0034&0.0034\\
\cline{3-8}
&No. of Features&\multicolumn{3}{M{30mm}|}{6000}&\multicolumn{3}{M{30mm}|}{17408 }\\
\hline 
\multirow{4}{*}{HC-F + Deep-F}&F1 &0.9854&0.9553&0.9533&0.9881&0.9244&0.9256\\
\cline{3-8}
&AUC&0.9994&0.9962&0.9962&0.9995&0.9912&0.9914\\
\cline{3-8}
&Loss&0.0006&0.0016&0.0016&0.0005&0.0034&0.0034\\
\cline{3-8}
&No. of Features&\multicolumn{3}{M{30mm}|}{6783}&\multicolumn{3}{M{30mm}|}{20039 }\\
\hline 
\multirow{4}{*}{HC-F + ResNet50} &F1 &0.9530&0.9575&0.9571&0.9625&0.9661&0.9674\\
\cline{3-8}
&AUC&0.9943&0.9958&0.9955&0.9966&0.9974&0.9974\\
\cline{3-8}
&Loss&0.0019&0.0014&0.0017&0.0014&0.0013&0.0013\\
\cline{3-8}
&No. of Features&\multicolumn{3}{M{30mm}|}{1783}&\multicolumn{3}{M{30mm}|}{4679 }\\
\hline
\multirow{4}{*}{\shortstack{HC-F +\\ DenseNet121}}  &F1 &0.9256&0.9286&0.9402&0.9593&0.9639&0.9661\\
\cline{3-8}
&AUC&0.9892&0.9912&0.9929&0.9955&0.9966&0.9970\\
\cline{3-8}
&Loss&0.0029&0.0027&0.0024&0.0016&0.0015&0.0014\\
\cline{3-8}
&No. of Features&\multicolumn{3}{M{30mm}|}{1783}&\multicolumn{3}{M{30mm}|}{3665 }\\
\hline
\multirow{4}{*}{\shortstack{HC-F +\\ InceptionV3}}&F1&0.9405&0.9432&0.9353&0.9564&0.9599&0.9582\\
\cline{3-8}
&AUC&0.9925&0.9923&0.9918&0.9960&0.9962&0.9959\\
\cline{3-8}
&Loss&0.0024&0.0023&0.0025&0.0016&0.0016&0.0016\\
\cline{3-8}
&No. of Features&\multicolumn{3}{M{30mm}|}{1783}&\multicolumn{3}{M{30mm}|}{4679 }\\
\hline \noalign{\smallskip}
\end{tabular}
\end{adjustbox}\\
\end{table}

We compared our best metric outputs with the study by korsuk et al. in \cite{sirinukunwattana2016locality}. They have used spatially constrained- CNNs (SC-CNN) to construct Softmax CNN + SSPP (Standard Single Patch Predictor) and Softmax CNN + NEP (Neighbouring Ensemble Predictor. They used their own customised CNN architecture with 2 convolution layers and 3 fully connected layers to classify nuclei patches. The output of the prediction is then evaluated using SSPP and NEP method which works on the principle that the pixel of interest is likely the centre of the nucleus and therefore, spatially constraining high probability pixels around the centre of the nucleus. The predicted output of the input patch $x_i$ is then weighted for its distance from the centred patch $x$, where $i$ is the ith patch in the set of the neighbouring patches scattered within radius $R$ of the patch $x$. The observations of their method on the nuclei dataset along with our ensemble deep learning methods are mentioned in Table \ref{tab9}. From the table, we observe that our deep ensemble method performs relatively better than both SCNN + SSPP and SCNN + NEP. Similarly, our ensemble handcrafted features method performs better than the complicated method proposed in \cite{sirinukunwattana2015novel,yuan2012quantitative} tested on the same dataset. Even, our cascaded ensemble HC-F performance comes close to SCNN + SSPP where the AUC value obtained by our model is ~2\% better than SCNN + SSPP and 0.38\% less than SCNN + NEP.  

\begin{table}[htbp]
\caption{Comparative performance measures with other comparative methods. The best performing metric is marked in bold}
\label{tab9}      
\centering
\renewcommand{\arraystretch}{1.5}
\begin{tabular}{|M{27mm}|M{23mm}|M{10mm}|P{10mm}|} 
\noalign{\smallskip} \hline
\multicolumn{2}{|c|}{Method} & F1-Score& AUC \\
\hline
\multirow{4}{*}{Deep methods}&\textbf{Deep ensemble} & \textbf{0.9874}&\textbf{0.9981} \\
\cline{2-4} 
&\textbf{HC-F + DL ensemble} &\textbf{0.9812}&\textbf{0.9978} \\
\cline{2-4} 
&SCNN + SSPP \cite{sirinukunwattana2016locality}&0.7480&0.8930 \\
\cline{2-4}
&SCNN + NEP \cite{sirinukunwattana2016locality} &0.7840&0.9170 \\
\hline
\multirow{3}{*}{handcrafted Methods}&\textbf{HC-F} &\textbf{0.7202}&\textbf{0.9132}  \\
\cline{2-4}
&CRImage \cite{yuan2012quantitative} &0.4880&0.6840  \\
\cline{2-4}
&Superpixel-Descriptor \cite{sirinukunwattana2015novel}&0.6870&0.8530  \\
\hline \noalign{\smallskip}
\end{tabular}
\end{table}
Table \ref{tab10} shows the comparison between deep fine-tuned models and ensemble models proposed in this paper. The obtained results summarizes that deep learning models alone may not perform well on the application; however, the performance of the ensemble models suggested that the combination of state-of-the-art deep and handcrafted models provides performance enhancement in case of colon cancer nuclei dataset. The performance gain with respect to CRCHistoPhenotype dataset is validated with another nuclei dataset in Section \ref{consep}. The results obtained in Table \ref{tab11} proves the statement that ensemble models perform better than single DL model.
\begin{table}[htbp]
\caption{Comparing fine-tuned deep learning models with ensemble models (CRCHistoPhenotypes)}
\label{tab10}      
\centering
\renewcommand{\arraystretch}{1.5}
\begin{tabular}{|M{27mm}|M{23mm}|P{10mm}|P{10mm}|P{10mm}|} 
\noalign{\smallskip} \hline
\multicolumn{2}{|c|}{\textbf{Method}} & \textbf{F1-Score} & \textbf{AUC} & \textbf{loss}\\
\hline
\multirow{4}{*}{Deep fine-tuned}&AlexNet & 0.8276  &0.9386 &1.1469  \\
\cline{2-5}
&VGG16 &  0.8689&0.9744 &0.8372  \\
\cline{2-5}
&VGG19 &0.8578&0.9701&0.9222 \\
\cline{2-5} 
&InceptionV3 &0.8175&0.9538&1.0795 \\
\cline{2-5}
&ResNet50 &0.8892&0.9799&0.7335 \\
\cline{2-5}
&DenseNet121 &0.8706&0.9756&0.7972  \\
\hline
\multirow{2}{*}{Ensemble 1}&Deep ensemble &0.9874&0.9981& 0.0130 \\
\cline{2-5}
&HC-F + DL ensemble &0.9812&0.9978& 0.0173 \\
\hline
\multirow{2}{*}{Ensemble 2}&Deep-F &0.9569&0.9965&0.0016  \\
\cline{2-5}
&HC-F + Deep+F &0.9553&0.9962&0.0016  \\
\hline \noalign{\smallskip}
\end{tabular}
\end{table}
\subsubsection{Performance validation with CoNSeP dataset}
\label{consep}
Colerectal Nuclear Segmentation and Phenotypes (CoNSeP) Dataset consists of 40 H\&E stained images of $1000 \times 1000$ dimensions. The dataset was recently published in  \cite{graham2019hover} for nuclei segmentation and classification problem. They provided the annotated class and instance maps as labels for each image. From the class maps, we extracted the class objects using standard image processing techniques and constructed a nuclei dataset for our study. The nuclei dataset comprise 7 classes namely, other, inflammatory, healthy epithelial, dysplastic/malignant epithelial, fibroblast, muscle, and endothelial. In \cite{graham2019hover} they combined the healthy and malignant epithelial nuclei into one class and spindle shaped nuclei - fibroblast, muscles, endothelial nuclei into one class. So, in total the four classes remained- epithelial, spindle-shaped, inflammatory, and Miscellaneous.  Each nuclei image extracted from the class map was kept $27\times 27$, same as we obtained from dataset in \cite{sirinukunwattana2016locality}. After extraction we obtained 1225 epithelial nuclei, 4543 spindle shaped nuclei, 1834 inflammatory, and 516 miscellaneous ones. As we can observe the dataset was highly imbalanced and hence to balance the dataset we followed the ADASYN algorithm from \cite{he2008adasyn}, same as we did for our first dataset \cite{sirinukunwattana2016locality}. Finally, we obtained the dataset comprising of total 18344 nuclei from CoNSeP dataset with balanced classes. \\
We performed the same set of experiments for this dataset and the table \ref{tab11} shows the results comparison between fine tuned deep models and ensemble models. We followed the comparison structure same as in table \ref{tab10}.  
\begin{table}[htbp]
\caption{Comparing fine-tuned deep learning models with ensemble models (ConSeP)}
\label{tab11}      
\centering
\renewcommand{\arraystretch}{1.5}
\begin{tabular}{|M{27mm}|M{23mm}|P{10mm}|P{10mm}|P{10mm}|} 
\noalign{\smallskip} \hline
\multicolumn{2}{|c|}{\textbf{Method}} & \textbf{F1-Score} & \textbf{AUC} & \textbf{loss}\\
\hline
\multirow{4}{*}{Deep fine-tuned}&AlexNet & 0.8848  &0.9790 &0.5343  \\
\cline{2-5}
&VGG16 &  0.9198&0.9873 &0.5343  \\
\cline{2-5}
&VGG19 &0.9035&0.9845&0.6005 \\
\cline{2-5} 
&InceptionV3 &0.8441&0.9600&0.8284 \\
\cline{2-5}
&ResNet50 &0.8495&0.9674&0.7262 \\
\cline{2-5}
&DenseNet121 &0.8156&0.9334&0.7743  \\
\hline
\multirow{2}{*}{Ensemble 1}&Deep ensemble &0.9753&0.9975& 0.0211 \\
\cline{2-5}
&HC-F + DL ensemble &0.9584&0.9959& 0.0314 \\
\hline
\multirow{2}{*}{Ensemble 2}&Deep-F &0.9786&0.9982&0.0187  \\
\cline{2-5}
&HC-F + Deep+F &0.9793&0.9980&0.0186  \\
\hline \noalign{\smallskip}
\end{tabular}
\end{table}

The experimental results show that the fine-tuned deep models perform relatively weak than ensemble models. The consistent results over two datasets validate our methodology. However, we noticed that lighter models like VGG16, VGG19, and AlexNet performed better than deeper models like ResNet, DenseNet, and InceptionV3. The lower performance metrics may be attributed to small image size ($27\times 27$) which needed to be resized to size $221\times 221$  since this is the minimum size required for input into these deep models. In contrast, for VGG16, VGG19, AlexNet, the images had to be resized to only $48\times 48$. Image resizing is firmly not recommended in the medical analysis since we lose a lot of high-resolution details. Hence, more massive deep models pose a bottleneck in case of small size images. Dataset balancing may have affected the performance of models, but validating this analysis remains out of the scope of this work. For future work, we could study the effect of image resizing and synthetic samples on the performance of the deeper models.

\subsubsection{Stability of the Experiments}   
We performed each experiment by randomly dividing our dataset into 70\% training, 15\% testing, and 15\% validation subsets. While the network improved on 70\% training and 15\% validation sets, the final results were obtained from an unseen test set when passed through trained models. We verified our ensemble models performance using 2, 5, and 10 fold cross-validation.  
\section{Discussion and Future Scope}

Histopathological data is highly variable and full of artefacts due to procedural faults during staining, fixing, and digitisation of biopsy samples. This phenomenon often occurs in all types of medical datasets and is generally more rampant in histopathological images. Also, due to heterogeneity among the same classes of cells and nuclei, it becomes difficult to generalise methods over a large number of samples. Nuclear morphology cannot be attributed to one kind of feature; its complex nature prevents the identification of all kinds of the cell with certainty. Moreover, unbalanced datasets that are prevalent and common in the field of biomedical images produce skewed performance and unintentionally cause the classifiers and feature learning process biased towards the class with a more significant number of samples. Therefore, the first step towards any CAD technique should be to acquire a balanced set of samples for even and unbiased learning. 
We therefore trained and tested our proposed methods on a balanced set.  Further, the experiments were designed to analyse models' performance. Keeping in mind the hypothesis we have taken up at the starting of this study, a thorough analysis has been put up in the form of the following observations. 
\begin{enumerate}
\item Individual handcrafted Features (HC-F) that extract features from raw images without object segmentation did not perform well on the classification task (Table \ref{tab4}). Except for LCOD and RSHD, all other local descriptors like LBP, LBDP, LDEP, LWP, HoG, and SURF could not encode colour information from the nuclei samples which may have been the main factor for their poor performance on our dataset. Whereas, in the case of BoVW, although it could encode colour information yet, it could not extract discriminative keypoint features because of the small sample dimensions. The nuclei patches were only of the size $27\times 27$, which did not allow the algorithm to declare multiple windows around the object to extract the key points. 
\item  Individual deep network output responses (Table \ref{tab5}) improved the classification performance to as much as 20\% relative to highest performing RSHD in handcrafted algorithms. These results helped determine that the generalising capacity of CNNs is far greater than domain-specific handcrafted features and that using an appropriate pre-trained model for fine-tuning is better than using handcrafted features. From the results \ref{tab5}, it was observed that F1-Scores of the networks lie within the range of 2 to 3\%, which tells that choosing which network would work well on the particular application could not be determined without tuning each model on different hyperparameters. Moreover,  choosing appropriate architecture also depends on the design choice, application platform, space, and time complexity. 
\item All finely tuned architectures were then used for extracting deep learning features. The features were then transferred to MLP layers to classify nuclei classes. This process of transfer learning significantly improved performance. The maximum range of improvement was ~11\% between Inceptionv3 in Table \ref{tab5} and InceptionV3 features in Table \ref{tab6}. These results, except VGG19, shows that a marked improvement can be achieved with transfer learning DL features instead of using end-to-end networks for classification. This statement also opens up a scope of further research in the domain to determine why transfer learning works better and what could be improved in end-to-end learning to achieve the same amount of performance. 
\item Observations in Table \ref{tab7} indicate that the cascaded ensemble of deep learning features performed better than using handcrafted features or deep learning features separately without cascading. The method of combining features also matters as experimentally proved through our second ensemble method. From Table \ref{tab8}, the recorded observations suggest that simple concatenation of features might not benefit the overall performance of the model and if proper feature engineering methods are not used, it might also cause either the decline in performance or no change.   The first ensemble method (cascaded ensemble) used PCA reduced features. In our case, not using dimensionality reduction with PCA would have caused increased space and time complexity. Hence, with feature reduction, we solved the problem of dimensionality and space constraints by a factor of ~34\%. In summary, we analysed through our experiments that in the case of deep cascaded networks using transfer learning methods to train extracted features, we could reduce the dimensions of the features using PCA without affecting the performance of the model.   
\item However, the most highlighting observation made in this study was that adding handcrafted features neither had any positive or negative effect on the model outcome besides small fluctuations in cross-entropy loss values (Table \ref{tab7} and \ref{tab8}). This might have happened due to the following reasons:
\begin{figure*} 
\centering
 \subfloat[short for lof][Handcrafted probability distribution\\for class 1]{
   \includegraphics[width=0.5\textwidth, height=1.5in]{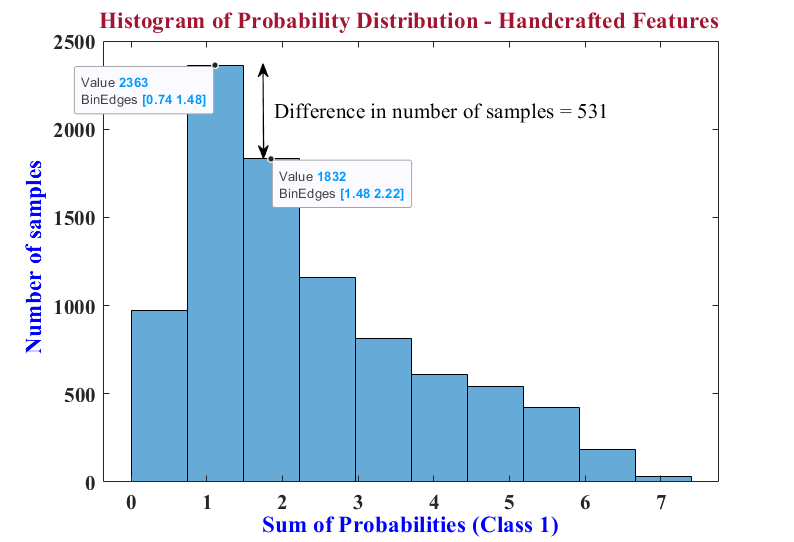}
   \label{subfig:fig1}
 }
 \subfloat[short for lof][Handcrafted probability distribution\\for class 2]{
   \includegraphics[width=0.5\textwidth, height=1.5in]{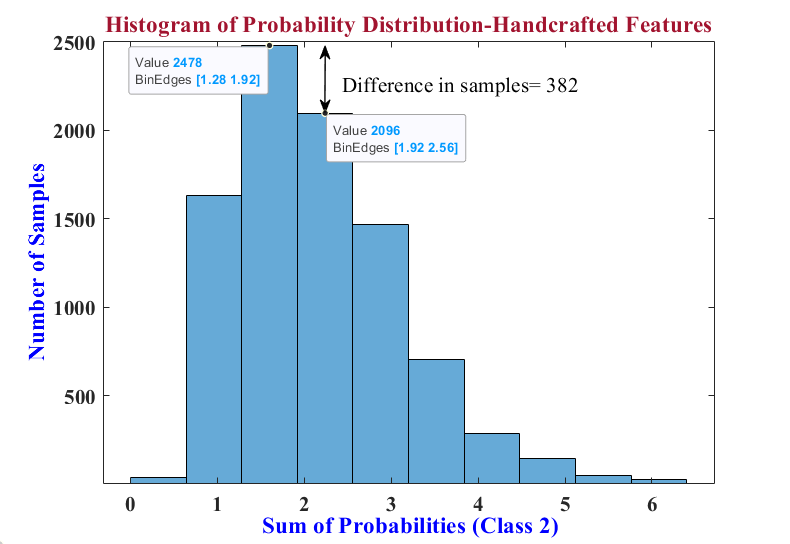}
   \label{subfig:fig2}
}\\
\subfloat[short for lof][Handcrafted probability distribution for\\class 3]{
   \includegraphics[width=0.5\textwidth, height=1.5in]{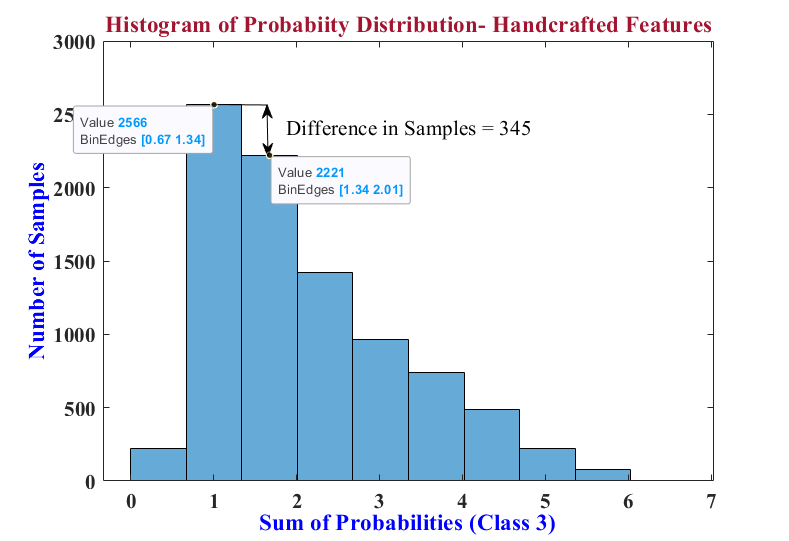}
   \label{subfig:fig3}
 }
 \subfloat[short for lof][Handcrafted probability distribution for\\class 4]{
   \includegraphics[width=0.5\textwidth, height=1.5in]{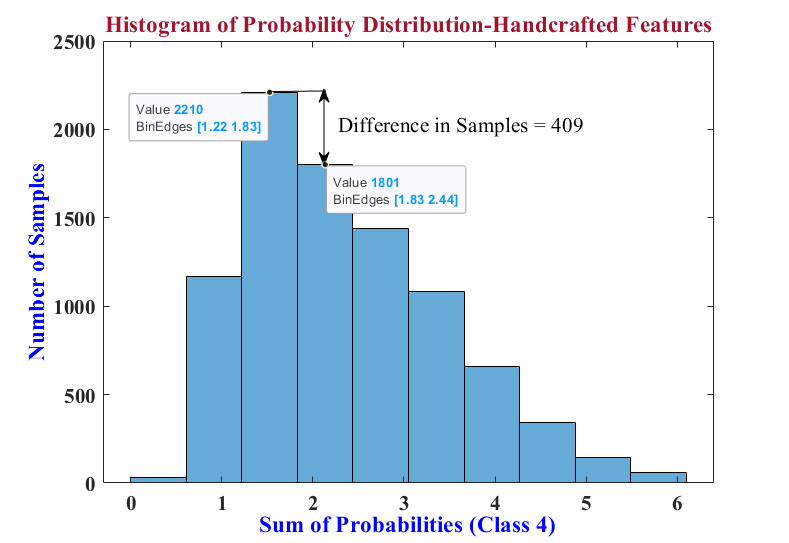}
   \label{subfig:fig4}
 }\\
\subfloat[short for lof][Deep Features probability distribution for\\class 1]{
   \includegraphics[width=0.5\textwidth, height=1.5in]{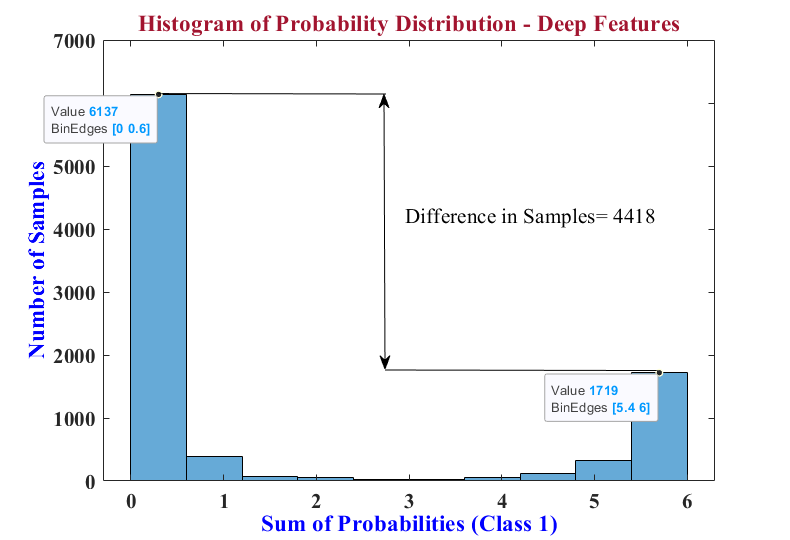}
   \label{subfig:fig5}
 }
 \subfloat[short for lof][Deep Features probability distribution for\\class 2]{
   \includegraphics[width=0.5\textwidth, height=1.5in]{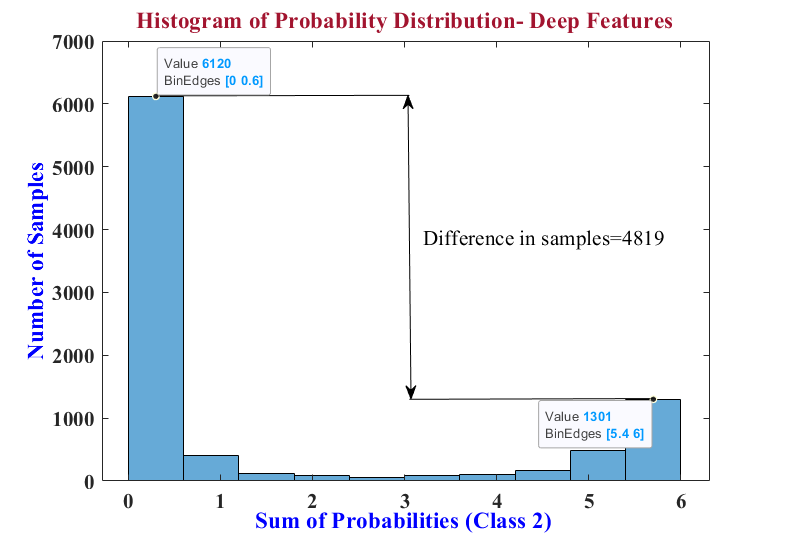}
   \label{subfig:fig6}
}\\
\subfloat[short for lof][Deep Features probability distribution for\\class 3]{
   \includegraphics[width=0.5\textwidth, height=1.5in]{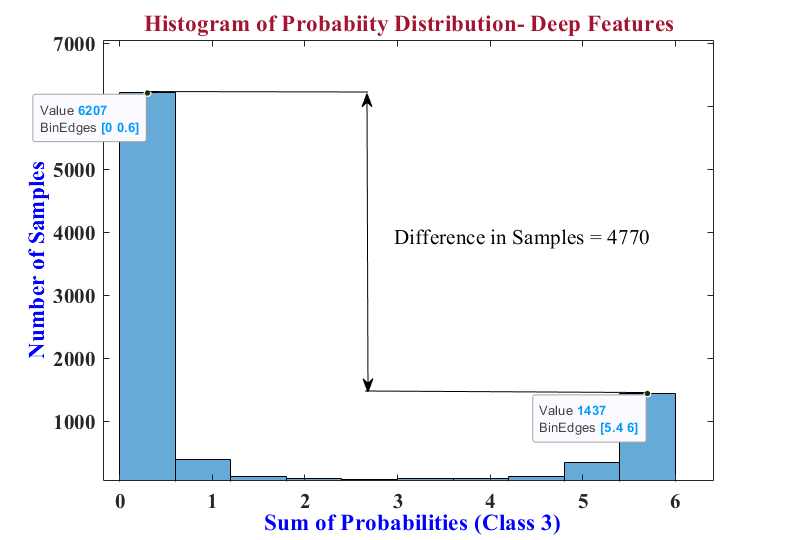}
   \label{subfig:fig7}
 }
 \subfloat[short for lof][Deep Features probability distribution for\\class 4]{
   \includegraphics[width=0.5\textwidth, height=1.5in]{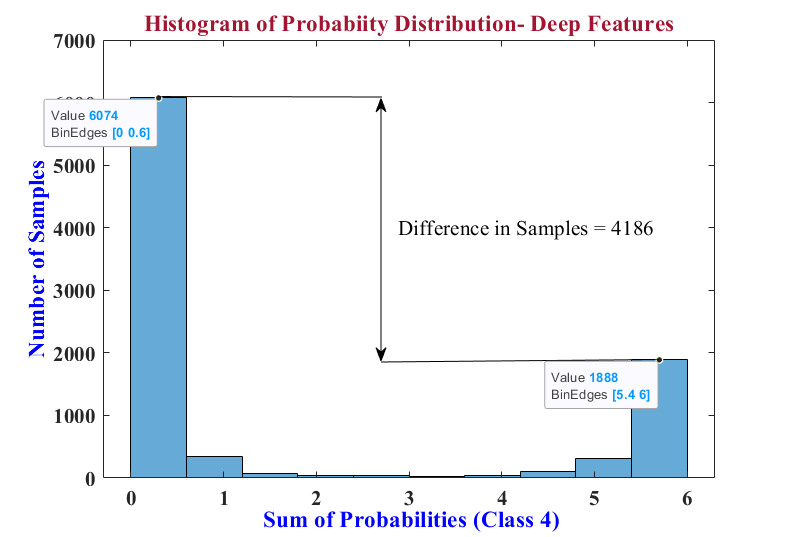}
   \label{subfig:fig8}
}\\
\caption[short for lof]{Histograms of probability distribution in case of Cascaded Ensemble method}
\end{figure*}
\begin{figure*} 
\ContinuedFloat
 \subfloat[short for lof][Combined Features probability\\ distribution for class 1]{
   \includegraphics[width=0.5\textwidth, height=1.5in]{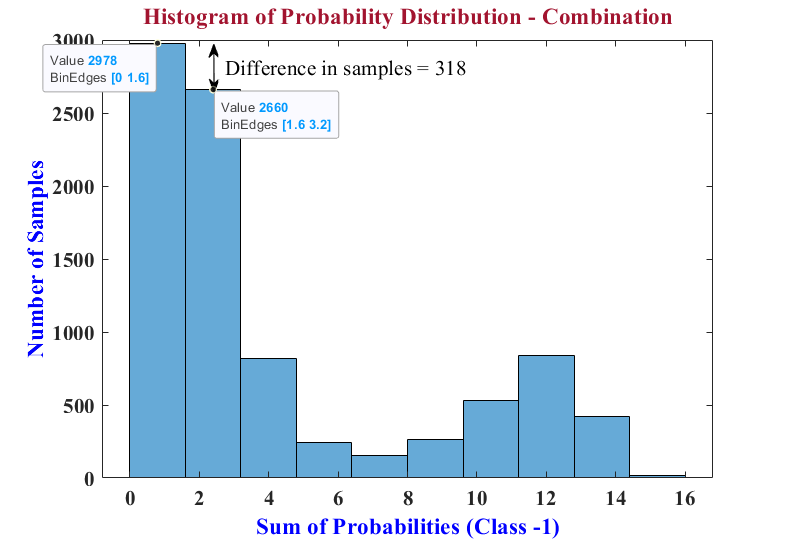}
   \label{subfig:fig9}
}
 \subfloat[short for lof][Combined Features probability\\ distribution for class 2]{
   \includegraphics[width=0.5\textwidth, height=1.5in]{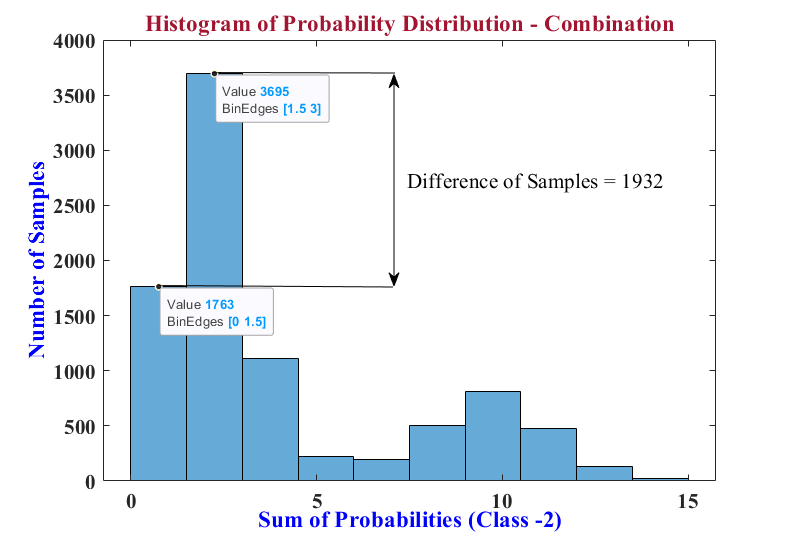}
   \label{subfig:fig10}
}\\
 \subfloat[short for lof][Combined Features probability\\ distribution for class 3]{
   \includegraphics[width=0.5\textwidth, height=1.5in]{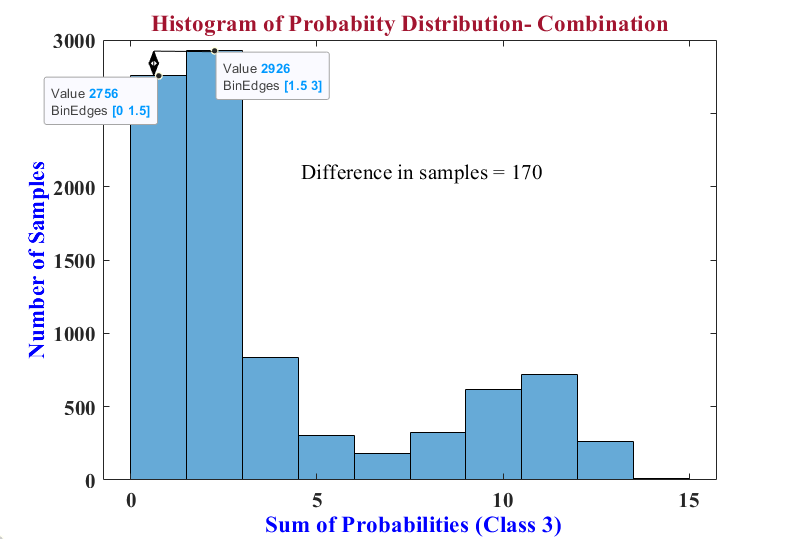}
   \label{subfig:fig11}
}
 \subfloat[short for lof][Combined Features probability\\ distribution for class 4]{
   \includegraphics[width=0.5\textwidth, height=1.5in]{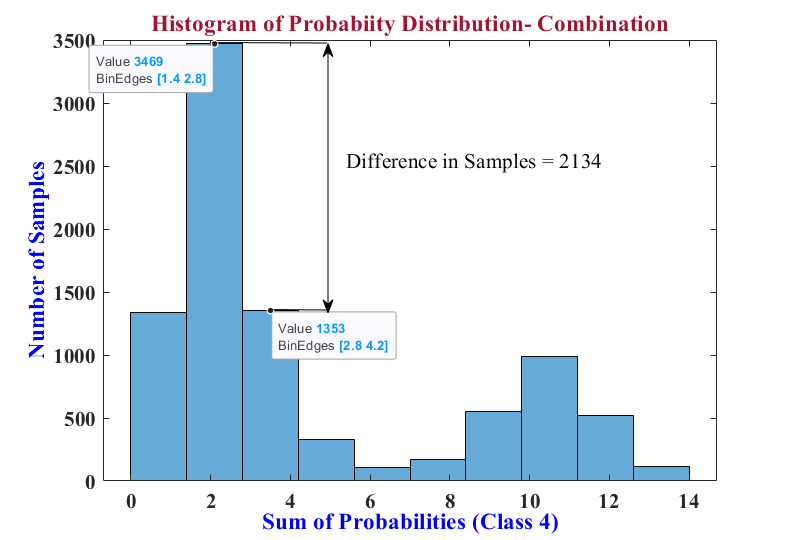}
   \label{subfig:fig12}
}
\caption[short for lof]{Histograms of probability distribution in case of Cascaded Ensemble method}
\label{fig5graphs}
\end{figure*}
\begin{itemize}
\item In the first method of ensembling features (Section \ref{CE}), when observation about the probabilities of two domains was made, it was found that the histogram of the probability distribution of handcrafted features shows less difference (for each class) between successive bins relative to the histogram of the probability distribution of deep features. To illustrate, we have plotted histograms of two methods for each class in Figure \ref{fig5graphs}. In the case of handcrafted features since the difference between probabilities is less for a large number of samples, when the two probabilities were summed, the difference in samples for combined distribution decreased due to the cumulative effect of both the distributions. The favourable distribution is in the case of deep features where the highest probability samples lie at the highest bin range and almost all rest of the samples lie at the lowest probability spectrum. This observation for the ensemble of deep features and ensemble of HC-F + Deep-F clarifies why there was no performance gain when handcrafted features were added to deep features. 
\item In the second method of ensembling features when features are concatenated along the column where each row represents one sample, a similar analysis can be made using the probability distribution curve. When handcrafted probabilities are concatenated, they bring no additional discriminative power to the deep features. Hence, there is no performance gain. 
\end{itemize} 
 
\end{enumerate}
This study focussed on recognising the effect of handcrafted features that were not extracted over the object but took the whole raw image as input, just as in case of deep learning methods. However, with complex datasets like histopathological nuclei images which suffer from challenges like low resolution, procedural artefacts, and high intra-class heterogeneity the discriminative power of feature descriptors were compromised. Hence they were discarded as a possibility for enhancing performance. Future research could focus on exploring ways to add more discriminative power with the help of better-handcrafted algorithms without the need for segmenting the objects.
\label{discussion}
\section{Conclusion}
Cell nuclei classification is among the widely followed disease diagnosis process that helps pathologists to segregate the type and grade of the cancer present in the biopsy sample of the patient. Recent approaches based on DL networks for classification alone could not be regarded as the ultimate solution for attaining efficient performance for all kinds of applications and neither adding more layers has any solid theoretical backup. Therefore, developing state of the art methods using existing models that have produced benchmark performance should be the focus area in current research. We have through this work, tried to analyse such methods that have recently gained momentum after proving their hypothesis that combining handcrafted and deep features result in performance gain in medical datasets. Through our experiments on two colon cancer nuclei datasets, we have observed that combining DL methods in different ways may prove to be a better approach than combining handcrafted features that are domain agnostic and work on raw images directly instead of extracting object-level features. The background and artefacts present around the nuclei create a deterrent which may have resulted in weak descriptor features preventing any significant addition to the already generalised and better performing deep features. Hence, we may need to analyse the type of features we want to extract from our images before combining them with deep features. Better combining techniques, along with suitable handcrafted and deep network candidates, could be explored to reduce computational and space complexity.
\section*{Acknowledgement}
This research was carried out in the Indian Institute of Information Technology, Allahabad and supported by the Ministry of Human Resource and Development, Government of India. We are also grateful to the NVIDIA corporation for supporting our research in this area by granting us TitanX (PASCAL) GPU.

\bibliographystyle{unsrt}
\bibliography{manuscript}

\end{document}